\documentclass[11pt]{article}
\usepackage[preprint]{acl}
\usepackage{times}
\usepackage{latexsym}
\usepackage[T1]{fontenc}
\usepackage[utf8]{inputenc}
\usepackage{microtype}
\usepackage{graphicx}
\usepackage{booktabs}
\usepackage{array}
\usepackage{makecell}
\usepackage{seqsplit}
\usepackage[hyphens]{xurl}
\usepackage{enumitem}
\setlist{nosep,leftmargin=*,topsep=2pt}
\newcommand{\noopsort}[1]{}
\newcommand{\plus}{$+$}
\newcommand{\minus}{$-$}

\title{GauntletScore: Auditable Fact-Checking That Matches the Best Published Checkers on LLM-AggreFact}

\author{Ernie Hobbs \quad Jim Hinson \\
  Genstrata Inc., Winston-Salem, NC \\
  Correspondence: \texttt{ernie.hobbs@genstrata.com}}

\makeatletter
\def\paragraph{\@startsection{paragraph}{4}{\z@}{1.5ex plus 0.5ex minus .2ex}{-0.33em}{\normalsize\bfseries}}
\makeatother

\begin{document}
\maketitle

\begin{abstract}
Checking whether a generated sentence is supported by its source document is a standard safeguard for AI-written text, and the strongest published checkers are single models. We describe GauntletScore, an evidence engine whose few rules and thresholds were set on the benchmark's development split. It puts typed questions to independent readers from different providers and records each answer as a measurement in an evidence ledger. It combines the measurements with a fixed Bayesian rule that counts a contradiction only when two readers agree. On the full LLM-AggreFact test split (29,320 claims across 11 datasets), following the leaderboard's approach, GauntletScore achieves 79.51 balanced accuracy, above every average published on the leaderboard as of 15 November 2026. That places it level with the best of them, nimble-v4-max (79.45 from its published dataset scores; the engine's 95\% interval, 78.47 to 80.53, contains it). On the same claims it beats two of its three readers used alone (Readers A and B) by 1.66 and 2.17 points (95\% intervals 0.80 to 2.53 and 1.29 to 3.06). Without AggreFact-CNN, described below, the leads are 0.90 and 1.32 and both intervals stay above zero. The engine also records the evidence behind every verdict, which a single model does not. Its largest margin is on full-length news articles checked against multi-sentence summaries (AggreFact-CNN), a format that resembles much regulated writing. There it scores 76.45, the best on the leaderboard, against 70.1 for the next best and 63.9 to 67.8 for the six systems with the highest averages. The test split was run once, on an engine frozen before the first test claim, under a stopping plan committed in advance. We release per-claim predictions and the scoring script.
\end{abstract}

\section{Introduction}

Even with the source document in hand, leading language models still add claims the source does not support. On Vectara's grounded-summarization leaderboard (22 October 2026), the newest models from OpenAI, Anthropic, Google and xAI had hallucination rates of 6.5 to 19.2 percent \citep{vectara2026}. Newer models do not consistently perform better: GPT-4.1, released in April 2025, scores 5.6 percent, below all four, and Gemini 4 Argon (12.1 percent) scores higher than Gemini 2.5 Pro (7.0 percent). Those errors have reached the courts, where a public database maintained by Damien Charlotin listed 2,292 cases involving AI-fabricated or suspect content in more than 40 countries as of November 2026 \citep{charlotin2026}. In one, a California appeals court sanctioned an attorney \$5,000 over a citation to a case that does not exist \citep{torrescampos2026}. In another, a federal court sanctioned three attorneys after eight of nine citations produced by their firm's AI tool proved fictitious \citep{wadsworth2025}. Courts are starting to require verification, and in September 2026 the Tenth Circuit proposed a rule that any filing prepared with generative AI first be reviewed by a human who verifies its cited authorities, quotations and citations \citep{tenthcircuit2026}.

The most dangerous errors sit in text that reads well. Fluent, well-organized text used to suggest that someone had done the work, and models now produce it without the work. In our own work, an AI-drafted outreach message credited an attorney with an appellate decision he had written about but never litigated. The sentence read naturally and went out unchecked, and our engine flagged it later. A safeguard that depends on a tired person remembering to check does not scale, so each claim has to be checked against its evidence every time.

Most automated checkers return a single model's judgment. Some are small models fine-tuned on synthetic examples of supported and unsupported claims (e.g., MiniCheck and its successors; \citealp{tang2024minicheck}), while others are general ``decision models'' that answer any yes-or-no question with a probability. Both kinds rely on one model's answer to a single question.

Our approach is closer to how people check facts. A claim is broken into parts, and each part is put to more than one independent reader as a typed question. Every answer is recorded as a measurement with its source, and a fixed Bayesian rule weighs the measurements, starting from even odds and capping how far any one answer can move the result. Evidence that shares an origin counts once, and a contradiction is accepted only when two readers agree. The output is a verdict level, with a record of every step behind it. By auditable we mean that each verdict can be traced to the stored answers and source passages that produced it.

This report tests that approach on LLM-AggreFact \citep{tang2024minicheck}, the public leaderboard for checking claims against a grounding document. Our contributions:

\begin{enumerate}
\item An evidence engine that scores 79.51 on the full test split, statistically indistinguishable from the best system published on the leaderboard (Section~\ref{sec:results}).
\item The best AggreFact-CNN score on the leaderboard is 76.45 on its full-length articles, which is ahead of the next best at 70.1 and the six systems with the highest averages at 63.9 to 67.8. This matters for regulated industries such as life sciences (Section~\ref{sec:analysis}).
\item A same-claims comparison with each reader used alone, reporting true-positive and true-negative rates beside the headline score (Sections~\ref{sec:results} and~\ref{sec:analysis}).
\item A protocol fixed in advance: the engine was frozen before the first test claim, the stopping plan was committed in advance, and every test claim ran exactly once (Section~\ref{sec:protocol}). We release per-claim predictions (Section~\ref{sec:repro}).
\item An accounting of where the design loses points, with each of the 29,320 verdicts traced to the stage that decided it, and of its limits (Sections~\ref{sec:analysis} and~\ref{sec:limits}).
\end{enumerate}

\section{Related work}

The engine draws on three lines of work: single fine-tuned checkers, the decision models we use as readers, and methods that combine several judges.

\paragraph{Benchmark:} LLM-AggreFact \citep{tang2024minicheck} collects human-labeled (document, claim, label) sets: AggreFact-CNN and AggreFact-XSum \citep{tang2023aggrefact}, TofuEval MediaSum and MeetingBank \citep{tang2024tofueval}, WiCE \citep{kamoi2023wice}, REVEAL \citep{jacovi2024reveal}, ClaimVerify \citep{liu2023verifiability}, FactCheck-GPT \citep{wang2024factcheckbench}, ExpertQA \citep{malaviya2024expertqa}, LFQA \citep{chen2024lfqa} and RAGTruth \citep{niu2024ragtruth}. The leaderboard scores balanced accuracy on each dataset and takes the unweighted mean of the eleven (leaderboard blog).\footnote{\url{https://llm-aggrefact.github.io/blog}}

\paragraph{Fine-tuned checkers:} Earlier checkers used entailment and question answering: FactCC \citep{kryscinski2020factcc}, SummaC \citep{laban2022summac}, QAFactEval \citep{fabbri2022qafacteval} and AlignScore \citep{zha2023alignscore}. MiniCheck \citep{tang2024minicheck} showed that a small model trained on synthetic data could match GPT-4 on this task. Bespoke-MiniCheck-7B (model card),\footnote{\url{https://huggingface.co/bespokelabs/Bespoke-MiniCheck-7B}} IBM's Granite Guardian \citep{padhi2025granite} and FactCG \citep{lei2025factcg} follow that line.

\paragraph{Decision models:} In 2026 several companies released models that answer typed questions with probabilities instead of generating text. TypeSafe's Jev, Cloudflare's Clef, Perplexity's Decider, Liquid AI's d1-ultra and Bespoke Labs' Nimble \citep{modelcards2026} now hold the leaderboard's top five. Decision models are also ranked on two community-run boards, the Jev Decision Index and JevBench \citep{jevindex2026,jevbench2026}. The Decision Index covers tools, classification, language, knowledge, and games and arts. JevBench, from Benchmark Heaven, covers typed decisions such as request routing, policy questions and intent classification. GauntletScore uses decision models as readers and is not one itself, so we did not enter either board. Only a few of the Index's 37 tasks, among them ANLI, ContractNLI and NLI4CT, check a statement against a document, whereas every LLM-AggreFact dataset does.

\paragraph{Combining judges:} The work on LLMs as judges \citep{zheng2023judging,liu2023geval} led to panels of judges \citep{verga2024juries} and to debate between models \citep{du2024debate}. Combining noisy labelers has a much longer history \citep{dawid1979}. Two recent preprints are closest to this work. RAIM \citep{onofri2026raim} combines ten open-weight judges of 4 to 9 billion parameters through a cross-fitted stacked logistic regression. It uses the judges' own outputs to test whether the panel will beat its best member. On eight faithfulness benchmarks it gave up about 2.9 balanced-accuracy points to a frontier judge, at about 1/64 of the inference price. Jev-as-a-Judge \citep{dorosario2026jev} escalates a claim from a decision model to a frontier model only when the decision model is not confident. On 495 LLM-AggreFact claims it found that the savings came from the share of claims escalated rather than from the confidence signal. It also found that the cascade changed cost more than accuracy and that the decision model and the frontier model were statistically tied in accuracy while making different errors. Our scorer's weights are fixed in advance rather than learned and we count evidence that shares an origin once and treat agreement and contradiction asymmetrically.

\paragraph{Evaluation metrics:} Two checkers can reach the same balanced accuracy with different errors \citep{godbole2025verify}. A headline accuracy can hide how many unsupported claims a checker catches \citep{dorosario2026jev}. A gap of a point or less between two checkers may not be informative, since our own 95\% interval is about two points wide. We therefore report confidence intervals, and true-positive and true-negative rates beside the headline score.

\section{System overview}
\label{sec:system}

GauntletScore turns a claim into a set of typed measurements and decides the verdict with a fixed Bayesian rule.

\begin{figure*}[t]
\centering
\includegraphics[width=0.92\textwidth]{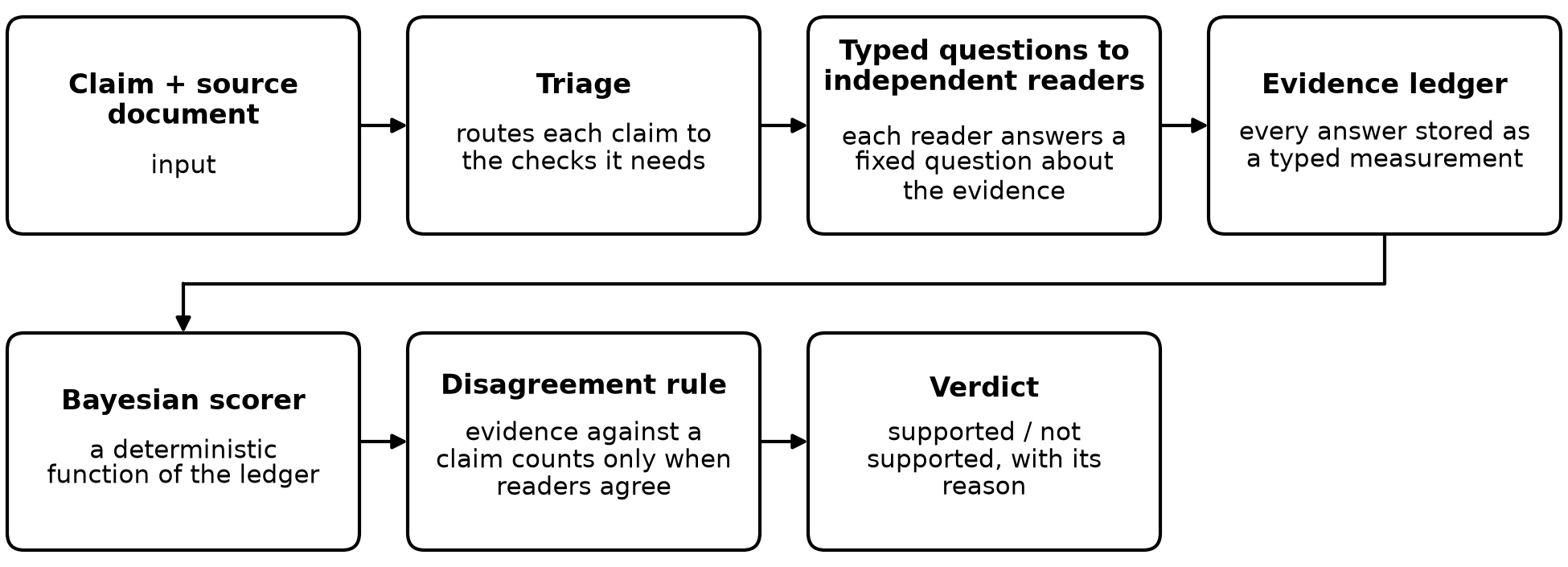}
\caption{The engine's stages, from a claim and its source document to a verdict with its reason.}
\label{fig:stages}
\end{figure*}

The engine runs in five stages (Figure~\ref{fig:stages}):

\begin{enumerate}
\item \textbf{Triage:} A fast reader works out what kind of claim it is and what has to be checked.
\item \textbf{Typed questions:} Each part of the claim is put to independent readers as a closed question with a fixed set of answers.
\item \textbf{Evidence ledger:} Every answer is stored as a typed measurement with its reader, its source and its strength. Every verdict can be traced back to the answers behind it, because all of them are kept.
\item \textbf{Bayesian scorer:} The scorer starts at even odds and applies a bounded likelihood ratio for each measurement. Evidence that shares an origin counts once. A partial answer moves the score less than a full one. The parts of a claim are combined so that the claim holds only if its parts hold.
\item \textbf{Disagreement rule and verdict:} A contradiction only counts if two readers agree. A third reader checks the claim unless the first two both find it supported with high confidence. If the readers disagree, a weighted vote decides. The result maps to a verdict level (Confirmed, Probably true, Leans true, and lower levels). For the benchmark, a claim counts as supported only at Confirmed.
\end{enumerate}

\paragraph{Post-scoring rules:} One compares the numbers in a claim with those in its source, and the other keeps a claim from being confirmed when both whole-claim readings doubt it. Together they decided 189 of the 29,320 test claims (Section~\ref{sec:analysis}).

\paragraph{Readers:} The run used three readers from three different providers: Readers A and B are commercial decision models, and Reader C is a general-purpose frontier model. The engine treats readers as interchangeable parts, and they can be swapped without retraining anything. This report tests one configuration.

\paragraph{Reader ordering:} The order is set by cost and speed. Readers A and B are fast, low-cost models. Both read every claim. Reader C is slower and costs more per call, so it is asked only when Readers A and B don't both find the claim supported with high confidence. On this benchmark Readers A and B settled 5,911 of the 29,320 claims this way (20\%), and their verdict was right on 5,837 of them (98.7\%). Reader C was still needed on about four claims in five, because the configuration for this run was tuned for accuracy. Raising the share of claims settled without Reader C is future work.

\paragraph{Benchmark mode:} In normal use the engine also searches the web and public records, weighs sources by tier, and checks causal claims. LLM-AggreFact supplies the grounding text for each claim, so for this run that text was the only evidence source. Retrieval, multi-source weighing and causal validation were switched off, so the benchmark tests only part of the engine.

\paragraph{Audit reports:} Outside of benchmark mode, each run produces a downloadable report. For every claim it gives the verdict level, it gives a short reason and the sources consulted (each with its tier, whether it supports or contradicts the claim, and the passage relied on). It also marks readings that a second reader did not confirm and lists statements judged to be opinions or forecasts without scoring them. No reports were used in this evaluation, but the analysis in Section~\ref{sec:analysis} draws on the same stored records.

\section{Evaluation protocol}
\label{sec:protocol}

The full test split was run once, on an engine frozen before the first test claim. It was scored by script against the human labels only.

\paragraph{Data:} The LLM-AggreFact test split from Hugging Face (lytang/LLM-AggreFact, revision 981dfd0b, SHA-256 \texttt{\seqsplit{ba6976c3da990a9d7b11670efaf4e3c93c1bd0530061ec742afd6b2d5bce698c}}, which matches the hash Hugging Face lists): 29,320 claims (Table~\ref{tab:data}).

\begin{table}[t]
\centering
\small
\begin{tabular}{lr}
\toprule
Dataset & Claims \\
\midrule
RAGTruth & 16,371 \\
ExpertQA & 3,702 \\
LFQA & 1,911 \\
REVEAL & 1,710 \\
FactCheck-GPT & 1,566 \\
ClaimVerify & 1,088 \\
TofuEval-MeetB & 772 \\
TofuEval-MediaS & 726 \\
AggreFact-CNN & 558 \\
AggreFact-XSum & 558 \\
WiCE & 358 \\
\bottomrule
\end{tabular}
\caption{Claims per dataset in the LLM-AggreFact test split.}
\label{tab:data}
\end{table}

\paragraph{Scoring:} Balanced accuracy (the mean of the true-positive and true-negative rates) on each dataset, then the unweighted mean of the eleven, as the leaderboard does. The engine outputs a binary label, and nothing was tuned per dataset or on the test split. GauntletScore refused 7 claims before any model call, so they are scored as not supported. On 10 more, a provider error left one of the two whole-claim readings missing, but the engine's own label stood and those claims were not rerun.

\paragraph{Development:} All development used the benchmark's development split, in 22 batches plus one separate 1,000-claim set: 7,086 of its 30,420 claims were drawn, and at most 6,206 ran through the engine. The test split was never used to choose a rule, prompt, threshold or model. No model was trained; the engine's rules and thresholds were chosen on the development split.

\paragraph{Test-set safeguards:}
\begin{itemize}
\item The stopping plan (four checkpoints, each with a futility line) was committed to version control before any test claim ran (commit 29b60eb). The run script refused to start unless that file was committed and unchanged.
\item The engine was frozen at one version (commit 682d0204) for the whole run. The reader lineup was read back at the start and every 500 claims.
\item Claims ran one at a time in a random order fixed by seed 20261108, with each checkpoint holding a random share of every dataset.
\item During the run only per-dataset and overall scores were looked at. No per-claim test result was read until the final number was locked.
\item The decision to run all four checkpoints, whatever checkpoint 3 scored, was recorded before checkpoint 3's result existed. Because the reported score covers every test claim, that decision could not pick a favorable stopping point.
\item The run script was changed twice during the run, each change committed on its own: a retry when the operating system briefly locked a file, and a running spend total in place of a recount. Neither touched the engine, the readers or the scoring.
\end{itemize}

\paragraph{Readers alone:} For every claim, the engine stores Reader A's and Reader B's own reading of the whole claim against the whole document. Counting a reading of 0.5 or more as supported gives each reader's score alone on exactly the same claims, at no extra cost.

\paragraph{Reader C alone:} Reader C makes no whole-claim reading inside the engine. That was a separate run after the test score was locked. It read each claim once against its whole document, with the same whole-claim instruction. The stratified sample of 2,992 test claims was drawn with a fixed seed before any call (272 per dataset, as near to half supported and half unsupported as each dataset allows). A probability of 0.5 or more counts as supported, the same rule as for Readers A and B. Reader C returned no readable answer for 13 claims, which count as not supported. Balanced accuracy does not depend on the label mix, so the sample scores each dataset fairly, with wider intervals than the full split.

\section{Results}
\label{sec:results}

GauntletScore scores 79.51 on the full test split, which is slightly better than the best published average of 79.45 (Table~\ref{tab:leaderboard}). Table~\ref{tab:readersAB} compares it with Readers A and B alone on the same claims. Table~\ref{tab:sample} and Appendix~\ref{app:readerC} compare it with Reader C alone on a sample.

\begin{table*}[t]
\centering
\footnotesize
\setlength{\tabcolsep}{3.4pt}
\resizebox{\textwidth}{!}{%
\begin{tabular}{llrrrrrrrrrrrr}
\toprule
Model & Size & Avg & CNN & XSum & MediaS & MeetB & WiCE & REVEAL & ClaimVerify & FactCheck & ExpertQA & LFQA & RAGTruth \\
\midrule
\makecell[l]{\textbf{GauntletScore}\\\textbf{(this work)}} & -- & \textbf{79.51} & 76.45 & 74.95 & 74.74 & 80.38 & 81.94 & 90.63 & 77.52 & 80.89 & 59.98 & 89.33 & 87.84 \\
\midrule
nimble-v4-max & 27B & 79.5 & 67.6 & 78.6 & 73.3 & 82.4 & 87.1 & 89.0 & 75.9 & 80.2 & 60.4 & 89.0 & 90.5 \\
Clef & 27B & 79.0 & 67.3 & 75.7 & 75.3 & 81.1 & 85.8 & 89.6 & 78.6 & 80.8 & 59.0 & 89.0 & 87.3 \\
Decider v1.2 & 27B & 78.7 & 65.5 & 77.9 & 72.4 & 80.3 & 87.1 & 90.3 & 75.8 & 78.7 & 60.6 & 88.4 & 89.0 \\
Jev 1.13.0 & -- & 78.7 & 67.2 & 75.5 & 74.5 & 83.7 & 83.9 & 90.9 & 76.9 & 77.0 & 59.9 & 89.2 & 87.0 \\
d1-ultra & -- & 78.5 & 63.9 & 77.6 & 74.4 & 79.9 & 82.7 & 91.6 & 76.8 & 79.0 & 60.2 & 87.9 & 89.2 \\
nimble-v4 & 9B & 77.8 & 67.8 & 78.5 & 72.5 & 79.7 & 85.1 & 88.6 & 71.4 & 80.1 & 60.1 & 82.7 & 88.9 \\
\bottomrule
\end{tabular}}
\caption{\textbf{LLM-AggreFact test split,} balanced accuracy (\%). Published rows read from the leaderboard (\url{https://llm-aggrefact.github.io/}) on 9 November and rechecked on 15 November 2026; the leaderboard shows one decimal. The eleven published scores for nimble-v4-max average 79.45. The published rows are the six systems with the highest averages.}
\label{tab:leaderboard}
\end{table*}

\begin{table*}[t]
\centering
\small
\begin{tabular}{lrrrrr}
\toprule
Dataset & Engine & Reader A alone & Reader B alone & Engine minus A & Engine minus B \\
\midrule
Average of 11 & 79.51 & 77.85 & 77.34 & \plus1.66 & \plus2.17 \\
\midrule
AggreFact-CNN & 76.45 & 67.12 & 65.77 & \plus9.33 & \plus10.68 \\
LFQA & 89.33 & 84.87 & 87.14 & \plus4.46 & \plus2.19 \\
ClaimVerify & 77.52 & 73.54 & 74.41 & \plus3.98 & \plus3.11 \\
REVEAL & 90.63 & 88.04 & 90.78 & \plus2.59 & \minus0.15 \\
TofuEval-MediaS & 74.74 & 72.37 & 69.98 & \plus2.37 & \plus4.76 \\
RAGTruth & 87.84 & 85.80 & 87.90 & \plus2.04 & \minus0.06 \\
FactCheck-GPT & 80.89 & 81.29 & 80.22 & \minus0.40 & \plus0.67 \\
TofuEval-MeetB & 80.38 & 80.97 & 78.97 & \minus0.59 & \plus1.41 \\
ExpertQA & 59.98 & 60.71 & 60.28 & \minus0.73 & \minus0.30 \\
AggreFact-XSum & 74.95 & 75.79 & 73.59 & \minus0.84 & \plus1.36 \\
WiCE & 81.94 & 85.84 & 81.70 & \minus3.90 & \plus0.24 \\
\bottomrule
\end{tabular}
\caption{\textbf{The engine against Readers A and B alone,} on the same claims, balanced accuracy (\%). Full test split, 29,320 claims. Intervals are under Statistical tests below.}
\label{tab:readersAB}
\end{table*}

Against Reader A, the engine wins on six datasets and loses on five (by less than one point on four of them). Without CNN, its lead over Reader A is 0.90 points, and over Reader B 1.32 points.

\paragraph{Statistical tests:} Paired bootstrap over claims within each dataset (10,000 resamples). The engine's average is 79.51 (95\% interval 78.47 to 80.53). This interval contains nimble-v4-max's 79.45, so the two are statistically indistinguishable. The engine's lead over Reader A is 1.66 (0.80 to 2.53), and over Reader B 2.17 (1.29 to 3.06). Without CNN it is 0.90 (0.14 to 1.66) and 1.32 (0.54 to 2.09). All four intervals stay above zero.

\paragraph{Interpreting reader-alone scores:} Each reader-alone score is our own single question to that model. It is possible that the model's maker or another user with a different prompt could score it higher. These gaps measure what the engine adds over the same readings and may overstate its lead over each model at its best. Against the strongest single models on the leaderboard the margin is smaller: 0.06 over nimble-v4-max and 0.5 over Clef. The same caution applies to Reader C alone below, which was read once with one instruction of ours.

\paragraph{Reader C alone:} The sample is 2,992 claims, about a tenth of the test split. On it, the engine scores 80.00 and Reader C alone 79.63, a gap of \plus0.37 points (95\% interval \minus0.66 to \plus1.37; paired bootstrap, 10,000 resamples). Without CNN, the gap is \plus0.19 (\minus0.84 to \plus1.22). Both intervals cross zero, so the difference is not statistically significant. Scores on part of the split also moved: the engine's running score on the test split was 79.03 at the first checkpoint (4,739 claims), rose to 79.89 by the third, and finished at 79.51. We therefore draw no conclusion about which of the two scores higher. Only a full-split run could settle it. See Appendix~\ref{app:readerC} for per-dataset scores.

\paragraph{Error profiles:} The readers differ more in the kind of mistake they make than in their scores (Table~\ref{tab:sample}). Readers A and B accept most claims (each keeps about 89\% of supported claims and catches about 67\% of unsupported ones), whereas Reader C alone is the strictest. The engine sits between them. Against Reader C alone, it wrongly rejects fewer supported claims (true-positive rate \plus2.7 points, 95\% interval \plus1.3 to \plus4.2) and catches fewer unsupported ones (true-negative rate \minus2.0, \minus3.5 to \minus0.6). The two effects cancel in balanced accuracy. This follows from the design, in which evidence against a claim counts only when two readers agree, making the engine slower than its strictest reader to call a claim unsupported.

\begin{table}[t]
\centering
\footnotesize
\setlength{\tabcolsep}{4pt}
\begin{tabular}{lrrr}
\toprule
System & \makecell[r]{Supported\\claims kept\\(true-positive\\rate)} & \makecell[r]{Unsupported\\claims caught\\(true-negative\\rate)} & \makecell[r]{Balanced\\accuracy} \\
\midrule
Engine & 82.5 & 77.5 & 80.00 \\
Reader A alone & 89.5 & 67.8 & 78.67 \\
Reader B alone & 89.0 & 66.6 & 77.81 \\
Reader C alone & 79.7 & 79.5 & 79.63 \\
\bottomrule
\end{tabular}
\caption{\textbf{Supported claims kept and unsupported claims caught,} 2,992-claim sample (\%). Means over the 11 datasets.}
\label{tab:sample}
\end{table}

\paragraph{Implications for review:} In review work, a true statement flagged as false costs an expert's time and teaches people to ignore the checker. The engine counts evidence against a claim only when two readers agree, so it wrongly rejects fewer true claims than Reader C alone, and the price is that it catches fewer unsupported ones (Table~\ref{tab:sample}). Its verdicts can also be audited. Each is built from stored readings of each part of the claim, which a reviewer can inspect, while a single model returns one label. No single model decides the verdict either. Reader C is one of three readers and can be outvoted (Section~\ref{sec:analysis}). Also any reader can be replaced without retraining. We plan to replace the hosted readers with models we train and host ourselves (Section~\ref{sec:conclusion}).

\section{Analysis}
\label{sec:analysis}

The gain is largest on the longest documents. The cost is highest on Wikipedia citations (WiCE). Against Readers A and B the engine catches unsupported claims that one whole-claim reading lets through, and it rejects some supported ones as a result.

\paragraph{Per-dataset gains:}
\begin{itemize}
\item \textbf{AggreFact-CNN:} This was the widest margin. The engine scores 76.45, which is the best CNN score on the leaderboard. The next best is 70.1 (FactCG-DeBERTa-L). The six systems in Table~\ref{tab:leaderboard} score 63.9 to 67.8. The engine is 9.3 points above Reader A on the same claims. The engine catches 33 of CNN's 57 unsupported claims, where Reader A catches 21. The engine got 15 CNN claims right that Reader A got wrong. Eight of them turned on reading a multi-sentence summary one sentence at a time to isolate a single wrong detail. Reader C alone also reads these articles well (75.0 on the sampled CNN claims, against the engine's 77.1 on the same claims). With only 57 unsupported claims, the engine's 95\% interval is wide (69.8 to 83.0).
\item \textbf{LFQA and ClaimVerify:} On these two sets of long-form answers to questions, the engine gains 4.5 and 4.0 points over Reader A. On LFQA the engine's 89.33 is the best on the leaderboard.
\item \textbf{TofuEval-MediaS:} The engine is 2.4 points above Reader A and 4.8 above Reader B.
\end{itemize}

\paragraph{Relevance to regulated documents:} CNN is a long source with a multi-sentence summary that is mostly right with one wrong detail. The errors the engine caught there that Reader A missed were mostly of that kind, two sides swapped, or a wrong time or quantity, or a writer the article never names. In life sciences, the same pattern appears when a clinical study report is summarized for a regulatory submission or for a plain-language summary of a trial. One wrong dose or treatment arm makes an otherwise accurate document misleading and, in a regulated filing, non-compliant. A checker that judges the whole summary at once tends to pass it, whereas the engine checks each sentence against the source and records which one failed. That's why its lead over Readers A and B is also largest on the longest documents (below). We plan to test this next on clinical-trial reports (see Conclusion).

\paragraph{Per-dataset losses:}
\begin{itemize}
\item \textbf{WiCE:} The engine is 3.9 points below Reader A and below every leading published system (82.7 to 87.1). The engine keeps 71\% of WiCE's supported claims where Reader A keeps 88\%. It trails Reader A at every document length. In a sample of 30 WiCE errors, the engine's own misses mostly came from strictness, rejecting a fair paraphrase or refusing to confirm a claim over a trivial missing detail. In 9 of the 16 sampled claims it wrongly rejected, Readers A and B alone both had it right.
\item \textbf{AggreFact-XSum:} The engine is 0.8 points below Reader A and below every leading published system (75.5 to 78.6). XSum claims come from one-sentence summaries that paraphrase heavily. The engine keeps 65\% of XSum's supported claims, against 77\% for Reader A.
\item \textbf{ExpertQA:} Every system finds this set hard. The leading published systems sit between 59.0 and 60.6. Our engine finished among them at 59.98.
\end{itemize}

\paragraph{True-positive and true-negative rates:} The balanced accuracy is the mean of two rates, the share of supported claims a checker keeps and the share of unsupported claims it catches. This engine catches more unsupported claims than Reader A on all 11 datasets, by 4.5 points (REVEAL) to 21 points (CNN). But it keeps fewer supported claims on 10 of 11, by 2.4 points (CNN) to 17 points (WiCE). Averaged over the datasets, the engine keeps 81.9\% of supported claims and catches 77.2\% of unsupported ones. Reader A alone keeps 89.1\% and catches 66.6\%, and Reader B alone 88.4\% and 66.3\% (Appendix~\ref{app:rates}).

\paragraph{Document length:} The engine has the smallest lead over Readers A and B on the shortest source documents (under one point below 200 words), and the largest on the longest (about three points at 800 words and over). However, this pattern doesn't hold everywhere. On ExpertQA's longer documents the engine trails Reader A, and on WiCE it trails at every length (Table~\ref{tab:length}).

\begin{table*}[t]
\centering
\small
\begin{tabular}{lrrrrrr}
\toprule
Document length (words) & Claims & Engine & Reader A alone & Reader B alone & Engine minus A & Engine minus B \\
\midrule
Under 200 & 8,524 & 80.99 & 80.20 & 80.49 & \plus0.8 & \plus0.5 \\
200 to 399 & 11,185 & 80.17 & 78.45 & 77.62 & \plus1.7 & \plus2.5 \\
400 to 799 & 6,056 & 75.75 & 74.17 & 73.91 & \plus1.6 & \plus1.8 \\
800 and over & 3,555 & 77.44 & 74.71 & 74.30 & \plus2.7 & \plus3.1 \\
\bottomrule
\end{tabular}
\caption{\textbf{Balanced accuracy by source-document length, full test split (\%):} Each cell is the mean over the datasets that have both labels in that band (8 or 9 of the 11).}
\label{tab:length}
\end{table*}

\paragraph{Decisions by stage:} Every reading is stored in the ledger, so each of the 29,320 verdicts can be traced to the stage that decided it, with no further model call. The counts by stage:
\begin{itemize}
\item \textbf{Claims settled without Reader C:} When Readers A and B agreed with confidence and we didn't ask Reader C (5,911 claims), the verdict was right 98.7\% of the time.
\item \textbf{Vote:} When Readers A and B together overruled Reader C (175 claims), they were right on 144.
\item \textbf{Rules:} The rule that compares numbers rejected 107 supported claims and caught 29 unsupported ones. A rule that keeps a claim from being confirmed when both whole-claim readings doubt it rejected 36 and caught 17. Together they decided 189 of the 29,320 claims. But the counts alone don't give each rule's net effect on balanced accuracy, since unsupported claims are the rarer label. Both rules were adopted on development-split evidence, and we report them as they ran, because a change made now could not be tested on this split.
\end{itemize}

\section{Limitations}
\label{sec:limits}

The problem is that the benchmark is one narrow job. It depends on commercial models we don't control.

Only checking against supplied text was tested. Retrieval, weighing several sources, and causal validation were off, and with only the supplied text for each claim the Bayesian scorer is only partly exercised, so these results do not cover the rest of the engine. The readers are hosted commercial models pinned to dated versions. Providers can change or retire them, so an exact rerun may not be possible later, and others cannot rerun the engine at all. We release per-claim predictions so the score can be checked (Section~\ref{sec:repro}).

The test labels are public. We cannot rule out that a reader saw them in training, and the same is true of every system on the leaderboard. The engine's rules were chosen on the development split and may suit these eleven datasets' style. They scored 80.07 on the development batches they were built on (in-sample), 79.26 on a fresh development batch, and 79.51 on test.

Small gaps on this benchmark are not meaningful. The top five published systems span one point. The leaderboard publishes no confidence intervals. Our own 95\% interval is 78.47 to 80.53, which is about two points wide and contains the best published average. Several datasets are small (WiCE has 358 claims, CNN has 57 unsupported ones). One number also hides different errors: two checkers can reach the same balanced accuracy with different mistakes \citep{godbole2025verify}. The engine and Reader C alone are an example, close in balanced accuracy on a sample, but leaning opposite ways (Section~\ref{sec:results}). Appendix~\ref{app:rates} gives true-positive and true-negative rates for each dataset.

Reader C alone was measured only on a sample, where the engine's lead of 0.37 points is not statistically significant (95\% interval \minus0.66 to \plus1.37). Reader C alone was read once, with one instruction of ours. The error samples in Section~\ref{sec:analysis} were read by AI agents and not checked by a person.

\section{Reproducibility}
\label{sec:repro}

The released files are enough to check our score, but not enough to rerun the engine.

Released at \url{https://huggingface.co/datasets/Gauntletscore/llm-aggrefact-predictions}:
\begin{itemize}
\item Per-claim predictions for all 29,320 test claims: claim ID and label only, with no claim or document text, to respect the dataset's terms.
\item The scoring script, which rebuilds our row of Table~\ref{tab:leaderboard} from those predictions and the public labels.
\item The dataset revision and hash, the run order with its seed, and the IDs of the 2,992 sampled claims.
\item The stopping plan as committed, and the run log of the test run, with spend figures, reader identities and other closed details blacked out and the redaction stated.
\item The engine version identifier used for the run.
\end{itemize}

Engine code, prompts, rules, thresholds, and reader identities are kept closed.

\section{Conclusion}
\label{sec:conclusion}

An evidence engine that weighs off-the-shelf readers under a fixed Bayesian rule can match the best published checkers on grounded factuality, and it keeps the readings behind every verdict. On this benchmark it beats two of its readers, and on full-length news articles checked against multi-sentence summaries, which is a format common in regulated writing, it has the best score on the leaderboard by more than 6 points.

We're taking this forward in three stages. The first is to show that the method works. This paper is one part of that. It measures the engine against the best published checkers on a public benchmark. The other part is a validation study we began in early 2026 and are now repeating on the current engine so that we can compare its verdicts with those of human evaluators. The second stage is to run it on models we control. We'll fine-tune open-weight models for the reader roles and host them ourselves so that we don't depend on outside providers and we can rerun the results exactly. The third is to test it where errors cost the most: long, data-dense documents in regulated industries such as the life sciences, including clinical-trial reports, where an unsupported claim can have regulatory or clinical consequences.

\section*{Disclosures and acknowledgments}

\paragraph{Competing interests:} Both authors are founders and shareholders of Genstrata Inc., which sells GauntletScore.

\paragraph{Funding:} Self-funded by Genstrata Inc.

\paragraph{Patents:} Patent pending.

\paragraph{Use of AI tools:} AI assistants were used to draft and edit text, write analysis scripts, and search the literature. Every number in this report was produced by scripts from the run records, and every figure and reference was checked against its source. The authors take full responsibility for the content.

\paragraph{Acknowledgments:} We thank the LLM-AggreFact maintainers and the authors of the eleven datasets.

{\small
\bibliography{references}

\begin{thebibliography}{32}
\providecommand{\natexlab}[1]{#1}

\bibitem[{Charlotin(2026)}]{charlotin2026}
Charlotin. 2026.
\newblock \href {https://www.damiencharlotin.com/hallucinations/} {{{AI}
  Hallucination Cases database, update of 12 November 2026}}.

\bibitem[{Chen et~al.(2024)}]{chen2024lfqa}
Chen et~al. 2024.
\newblock Understanding retrieval augmentation for long-form question
  answering.
\newblock In \emph{Proceedings of COLM 2024}.

\bibitem[{Dawid and Skene(1979)}]{dawid1979}
Dawid and Skene. 1979.
\newblock Maximum likelihood estimation of observer error-rates using the {EM}
  algorithm.
\newblock \emph{Journal of the Royal Statistical Society, Series C (Applied
  Statistics)}, 28(1):20--28.

\bibitem[{{Do Rosario} and Atef(2026)}]{dorosario2026jev}
{Do Rosario} and Atef. 2026.
\newblock {Jev-as-a-Judge} for {RAG} claim verification: Confidence-gated
  cascading between a non-generative verifier and a frontier {LLM}.
\newblock arXiv 2610.20634.

\bibitem[{Du et~al.(2024)}]{du2024debate}
Du et~al. 2024.
\newblock Improving factuality and reasoning in language models through
  multiagent debate.
\newblock In \emph{Proceedings of ICML 2024}.

\bibitem[{Fabbri et~al.(2022)}]{fabbri2022qafacteval}
Fabbri et~al. 2022.
\newblock {QAFactEval}: Improved {QA}-based factual consistency evaluation for
  summarization.
\newblock In \emph{Proceedings of NAACL 2022}.

\bibitem[{Godbole and Jia(2025)}]{godbole2025verify}
Godbole and Jia. 2025.
\newblock \href {https://aclanthology.org/2025.findings-acl.1175/} {Verify with
  caution: The pitfalls of relying on imperfect factuality metrics}.
\newblock In \emph{Findings of ACL 2025}.

\bibitem[{Jacovi et~al.(2024)}]{jacovi2024reveal}
Jacovi et~al. 2024.
\newblock A chain-of-thought is as strong as its weakest link: A benchmark for
  verifiers of reasoning chains ({REVEAL}).
\newblock In \emph{Proceedings of ACL 2024}.

\bibitem[{{Jev Decision Index}(2026)}]{jevindex2026}
{Jev Decision Index}. 2026.
\newblock \href
  {https://huggingface.co/spaces/multimodalart/jev-decision-index} {{Jev
  Decision Index, edition 0.4.1, 21 October 2026. Community leaderboard for
  decision models, not affiliated with {TypeSafe}}}.

\bibitem[{{JevBench}(2026)}]{jevbench2026}
{JevBench}. 2026.
\newblock \href {https://github.com/fstandhartinger/jevbench} {{{JevBench}.
  {Benchmark Heaven}}}.

\bibitem[{Kamoi et~al.(2023)}]{kamoi2023wice}
Kamoi et~al. 2023.
\newblock {WiCE}: Real-world entailment for claims in {Wikipedia}.
\newblock In \emph{Proceedings of EMNLP 2023}.

\bibitem[{Kry{\'s}ci{\'n}ski et~al.(2020)}]{kryscinski2020factcc}
Kry{\'s}ci{\'n}ski et~al. 2020.
\newblock Evaluating the factual consistency of abstractive text summarization
  ({FactCC}).
\newblock In \emph{Proceedings of EMNLP 2020}.

\bibitem[{Laban et~al.(2022)}]{laban2022summac}
Laban et~al. 2022.
\newblock {SummaC}: Re-visiting {NLI}-based models for inconsistency detection
  in summarization.
\newblock \emph{Transactions of the Association for Computational Linguistics}.

\bibitem[{Lei et~al.(2025)}]{lei2025factcg}
Lei et~al. 2025.
\newblock {FactCG}: Enhancing fact checkers with graph-based multi-hop data.
\newblock In \emph{Proceedings of NAACL 2025}.

\bibitem[{Liu et~al.(2023{\natexlab{a}})Liu, Zhang, and
  Liang}]{liu2023verifiability}
Liu, Zhang, and Liang. 2023{\natexlab{a}}.
\newblock Evaluating verifiability in generative search engines.
\newblock In \emph{Findings of EMNLP 2023}.

\bibitem[{Liu et~al.(2023{\natexlab{b}})}]{liu2023geval}
Liu et~al. 2023{\natexlab{b}}.
\newblock {G-Eval}: {NLG} evaluation using {GPT-4} with better human alignment.
\newblock In \emph{Proceedings of EMNLP 2023}.

\bibitem[{Malaviya et~al.(2024)}]{malaviya2024expertqa}
Malaviya et~al. 2024.
\newblock {ExpertQA}: Expert-curated questions and attributed answers.
\newblock In \emph{Proceedings of NAACL 2024}.

\bibitem[{{Model cards}(2026)}]{modelcards2026}
{Model cards}. 2026.
\newblock {Model cards and release pages: Clef,
  \url{https://huggingface.co/Cloudflare/clef}; Decider v1.2,
  \url{https://huggingface.co/perplexity-ai/pplx-decider-v1.2}; d1-ultra,
  \url{https://www.liquid.ai/blog/d1-ultra}; Jev 1.13.0,
  \url{https://docs.typesafe.ai/primitives/choice}; nimble-v4-max,
  \url{https://huggingface.co/bespokelabs/Nimble-v4-max}}.

\bibitem[{Niu et~al.(2024)}]{niu2024ragtruth}
Niu et~al. 2024.
\newblock {RAGTruth}: A hallucination corpus for developing trustworthy
  retrieval-augmented language models.
\newblock In \emph{Proceedings of ACL 2024}.

\bibitem[{Onofri and {Di Pietro}(2026)}]{onofri2026raim}
Onofri and {Di Pietro}. 2026.
\newblock {RAIM}: Robust aggregation of inexpensive models for hallucination
  detection.
\newblock arXiv 2609.39229, 30 September 2026.

\bibitem[{Padhi et~al.(2025)}]{padhi2025granite}
Padhi et~al. 2025.
\newblock {Granite Guardian}: Comprehensive {LLM} safeguarding.
\newblock In \emph{Proceedings of NAACL 2025, Industry Track}.

\bibitem[{Tang et~al.(2024{\natexlab{a}})Tang, Laban, and
  Durrett}]{tang2024minicheck}
Tang, Laban, and Durrett. 2024{\natexlab{a}}.
\newblock {MiniCheck}: Efficient fact-checking of {LLMs} on grounding
  documents.
\newblock In \emph{Proceedings of EMNLP 2024}.

\bibitem[{Tang et~al.(2023)}]{tang2023aggrefact}
Tang et~al. 2023.
\newblock Understanding factual errors in summarization: Errors, summarizers,
  datasets, error detectors.
\newblock In \emph{Proceedings of ACL 2023}.

\bibitem[{Tang et~al.(2024{\natexlab{b}})}]{tang2024tofueval}
Tang et~al. 2024{\natexlab{b}}.
\newblock {TofuEval}: Evaluating hallucinations of {LLMs} on topic-focused
  dialogue summarization.
\newblock In \emph{Proceedings of NAACL 2024}.

\bibitem[{{Tenth Circuit}(2026)}]{tenthcircuit2026}
{Tenth Circuit}. 2026.
\newblock {U.S. Court of Appeals for the Tenth Circuit, Clerk of Court.
  Memorandum on 2027 revisions to the Tenth Circuit Rules, proposed 10th Cir.
  R. 46.5(B)(4), 18 September 2026}.

\bibitem[{{Torres Campos}(2026)}]{torrescampos2026}
{Torres Campos}. 2026.
\newblock \href
  {https://law.justia.com/cases/california/court-of-appeal/2026/d085584m.html}
  {{In re the Domestic Partnership of {Torres Campos} and {Munoz} (2026) 118
  Cal.App.5th 1112, Cal. Ct. App., 4th Dist., Div. 1, No. D085584 (5 March
  2026)}}.

\bibitem[{{Vectara}(2026)}]{vectara2026}
{Vectara}. 2026.
\newblock \href {https://github.com/vectara/hallucination-leaderboard}
  {{Hallucination leaderboard ({HHEM}), table of 22 October 2026}}.

\bibitem[{Verga et~al.(2024)}]{verga2024juries}
Verga et~al. 2024.
\newblock Replacing judges with juries: Evaluating {LLM} generations with a
  panel of diverse models.
\newblock arXiv 2404.18796.

\bibitem[{{Wadsworth v. Walmart}(2025)}]{wadsworth2025}
{Wadsworth v. Walmart}. 2025.
\newblock \href
  {https://www.courtlistener.com/docket/67564066/wadsworth-v-walmart-inc/}
  {{Wadsworth v. Walmart Inc., D. Wyo., No. 2:23-cv-00118-KHR, sanctions order
  of 24 February 2025}}.

\bibitem[{Wang et~al.(2024)}]{wang2024factcheckbench}
Wang et~al. 2024.
\newblock {Factcheck-Bench}: Fine-grained evaluation benchmark for automatic
  fact-checkers.
\newblock In \emph{Findings of EMNLP 2024}.

\bibitem[{Zha et~al.(2023)}]{zha2023alignscore}
Zha et~al. 2023.
\newblock {AlignScore}: Evaluating factual consistency with a unified alignment
  function.
\newblock In \emph{Proceedings of ACL 2023}.

\bibitem[{Zheng et~al.(2023)}]{zheng2023judging}
Zheng et~al. 2023.
\newblock Judging {LLM}-as-a-judge with {MT-Bench} and {Chatbot Arena}.
\newblock In \emph{NeurIPS 2023, Datasets and Benchmarks Track}.

\end{thebibliography}
}

\clearpage
\onecolumn
\appendix

\section{True-positive and true-negative rates by dataset}
\label{app:rates}

\begin{table}[h]
\centering
\small
\begin{tabular}{lrrrrrrrr}
\toprule
 & & & \multicolumn{3}{c}{Supported claims kept} & \multicolumn{3}{c}{Unsupported claims caught} \\
\cmidrule(lr){4-6}\cmidrule(lr){7-9}
Dataset & Supported & Unsupported & Engine & A alone & B alone & Engine & A alone & B alone \\
\midrule
AggreFact-CNN & 501 & 57 & 95.0 & 97.4 & 98.2 & 57.9 & 36.8 & 33.3 \\
AggreFact-XSum & 285 & 273 & 64.9 & 76.5 & 84.9 & 85.0 & 75.1 & 62.3 \\
ClaimVerify & 789 & 299 & 85.8 & 91.9 & 90.6 & 69.2 & 55.2 & 58.2 \\
ExpertQA & 2,971 & 731 & 60.5 & 70.1 & 63.2 & 59.5 & 51.3 & 57.3 \\
FactCheck-GPT & 376 & 1,190 & 72.9 & 78.7 & 71.3 & 88.9 & 83.9 & 89.2 \\
LFQA & 1,121 & 790 & 89.9 & 94.6 & 93.4 & 88.7 & 75.2 & 80.9 \\
RAGTruth & 15,102 & 1,269 & 84.9 & 92.7 & 92.5 & 90.8 & 78.9 & 83.3 \\
REVEAL & 400 & 1,310 & 96.0 & 95.2 & 97.8 & 85.3 & 80.8 & 83.8 \\
TofuEval-MediaS & 554 & 172 & 91.3 & 97.7 & 96.9 & 58.1 & 47.1 & 43.0 \\
TofuEval-MeetB & 622 & 150 & 88.1 & 97.3 & 97.3 & 72.7 & 64.7 & 60.7 \\
WiCE & 111 & 247 & 71.2 & 88.3 & 86.5 & 92.7 & 83.4 & 76.9 \\
\midrule
Mean of 11 & 22,832 & 6,488 & 81.9 & 89.1 & 88.4 & 77.2 & 66.6 & 66.3 \\
\bottomrule
\end{tabular}
\caption{Supported claims kept (true-positive rate) and unsupported claims caught (true-negative rate), full test split (\%). The last row gives total counts and mean rates.}
\label{tab:rates}
\end{table}

\section{The engine and Reader C alone by dataset}
\label{app:readerC}

\begin{table}[h]
\centering
\small
\begin{tabular}{lrrr}
\toprule
Dataset & Engine & \makecell[r]{Reader C\\alone} & \makecell[r]{Engine\\minus C} \\
\midrule
Average of 11 & 80.00 & 79.63 & \plus0.37 \\
\midrule
AggreFact-CNN & 77.09 & 74.99 & \plus2.10 \\
AggreFact-XSum & 74.26 & 76.47 & \minus2.21 \\
ClaimVerify & 77.21 & 76.10 & \plus1.11 \\
ExpertQA & 63.24 & 60.29 & \plus2.95 \\
FactCheck-GPT & 79.78 & 78.31 & \plus1.47 \\
LFQA & 91.18 & 90.44 & \plus0.74 \\
RAGTruth & 88.24 & 88.60 & \minus0.36 \\
REVEAL & 91.54 & 91.18 & \plus0.36 \\
TofuEval-MediaS & 75.00 & 76.84 & \minus1.84 \\
TofuEval-MeetB & 80.88 & 81.99 & \minus1.11 \\
WiCE & 81.55 & 80.73 & \plus0.82 \\
\bottomrule
\end{tabular}
\caption{The engine and Reader C alone on the 2,992-claim sample, balanced accuracy (\%). 272 claims per dataset, so single-dataset gaps of 2 to 3 points are within noise. Readers A and B on the full split are in Table~\ref{tab:readersAB}.}
\label{tab:readerC}
\end{table}

\end{document}